\def\year{2020}\relax
\documentclass[letterpaper]{article} 
\usepackage{aaai20}  
\usepackage{times}  
\usepackage{helvet} 
\usepackage{courier}  
\usepackage[hyphens]{url}  
\usepackage{graphicx} 
\usepackage{graphicx}  
\usepackage{xcolor}

\nocopyright

\title{Accelerating the Development of Multimodal, Integrative-AI Systems with Platform for Situated Intelligence}
\author{Sean Andrist, Dan Bohus\\ 
Microsoft Research\\ 
\{sandrist;dbohus\}@microsoft.com 
}

\newcommand{\psif}{\emph{\textbackslash psi} }
\newcommand{\psifnospace}{\emph{\textbackslash psi}}

\begin{document}

\maketitle

\begin{abstract}
We describe Platform for Situated Intelligence, an open-source framework for multimodal, integrative-AI systems. The framework provides infrastructure, tools, and components that enable and accelerate the development of applications that process multimodal streams of data and in which timing is critical. The framework is particularly well-suited for developing physically situated interactive systems that perceive and reason about their surroundings in order to better interact with people, such as social robots, virtual assistants, smart meeting rooms, etc. In this paper, we provide a brief, high-level overview of the framework and its main affordances, and discuss its implications for HRI.
\end{abstract}

\section{Introduction}

Recent years have brought renewed excitement and interest in the possibility of AI-infused robots that can interact with people in the open world. Unfortunately, the barrier-to-entry for conducting research in this space is still very high, as anyone who has attempted to build such systems can attest. Before reaching the interesting research and design challenges, one must often first overcome the daunting engineering tasks involved in building end-to-end prototype systems that serve as test-beds for experimentation. Due to these challenges, and the large impedance mismatch with respect to the generic software development infrastructures and tools available today, it is not surprising that progress in this area has been slow, despite advances in individual component technologies.

The last decade has indeed yielded fast-paced progress and significant breakthroughs in several individual areas of AI and machine learning, from natural language processing models that generate believable text in nearly any domain \cite{brown2020language}, to computer vision systems for accurate object detection and pose estimation \cite{Detectron2018}, and robot control algorithms for dexterous manipulation \cite{andrychowicz2020learning}. These breakthroughs have been fueled in part by large datasets, increased computing power, new techniques for faster training of deep-learned models, and improvements in sim2real transfer \cite{kadian2019we}. However, combining \emph{multiple} such AI technologies into end-to-end intelligent systems remains very challenging and time-consuming today.

Robots that understand and interact with people in the open world are a prime example of a general class of \emph{multimodal, integrative-AI systems}. These systems need to weave together and carefully orchestrate a heterogeneous set of AI components and sensors which generate and process data across different modalities, such as audio, video, depth, speech, lidar, etc. Often they must operate under strict latency constraints, and components must execute asynchronously for efficiency reasons, yet they must also be closely coordinated in real time and finely tuned in order to create a well-functioning end-to-end system. Unfortunately, the typical programming languages used in developing these systems lack important primitives and features for reasoning about time and latency. Without support for operations like synchronization and data fusion, a significant amount of development time is spent debugging low-level problems, rather than on the high-level tasks at hand. Researchers are forced to constantly ``reinvent the wheel,'' building their own custom infrastructure for representing and reasoning about important timing constructs, which often ends up serving only the needs of their specific application, hindering generalization and reuse.

Furthermore, multimodal, integrative-AI applications have a specific set of debugging, visualization, and analytics needs that are not sufficiently addressed by the development tools generally available today. Standard debugging techniques such as the use of breakpoints in code or ad-hoc ``printf debugging'' are insufficient. Instead, the ability to visualize the data as it actually flows through the application is paramount. Formulating analyses and queries over the data further accelerates the development and tuning process.

To address these challenges and alleviate the high engineering costs, our team has developed Platform for Situated Intelligence\footnote{\url{https://github.com/microsoft/psi}} (in short \psifnospace, pronounced like the Greek letter $\psi$), an open-source  and extensible framework to support more rapid development and research in multimodal, integrative-AI systems, such as socially interactive autonomous robots. The framework is intended to simplify the development, debugging, analysis, maintenance, and continuous evolution of such systems by empowering rapid prototyping and iteration.

The development of \psif was motivated and informed by our team's experiences in the robotics space, as well as research on how to create physically situated, multimodal interactive systems that understand social context and fluidly interact with groups of people \cite{bohus2009models,bohus2011,andrist2016you,bohus2017study,tan2020now}. This research has involved the development of fully autonomous prototype systems that are deployed in the open world for extended periods, including robots that give directions to people in our building \cite{bohus2014directions}. Throughout this research, as well as in our broader interactions with the robotics, HRI, and multimodal interaction research communities, we have experienced and observed many of the challenges listed above. In the next section, we present a brief high-level architectural overview of \psifnospace. For the interested reader, a longer in-depth description of the framework's various affordances is available in \cite{bohus2020psi}.

\section{Overview}

Platform for Situated Intelligence is comprised of three aspects: (1) the \textit{runtime} infrastructure that provides support for working with temporal streams of data, and a model for parallel coordinated computation; (2) a set of \textit{tools} that enable debugging, multimodal data visualization, annotation, and processing; and (3) an open, growing ecosystem of \textit{components} that wrap various AI technologies and sensors.

At a high level, an application built with \psif can be broken down into a few basic concepts. A \psif application is basically a graph, or \emph{pipeline}, of nodes and edges. Each node represents a \emph{component} that encapsulates a specific AI technology, sensor, effector, or other unit of computation. Examples in the vision domain include components for pose tracking, cameras, rendering an image to a screen, and image cropping respectively. Each edge in the pipeline graph represents a one-way stream of data messages of a particular type. For example, the stream exiting a camera component carries image messages. Streams of a specific type can only be connected to components that receive the same data type.

Pipelines are easily constructed by instantiating components and connecting them together with imperative code. Once the pipeline is started, the \psif \emph{runtime} controls its execution, and in the process provides a number of affordances.

\subsection{Runtime Infrastructure}

The \psif runtime infrastructure provides a programming and execution model for parallel, coordinated computation based on time-aware data streams. This infrastructure is built on .NET Standard to be cross-platform. It retains the affordances and software engineering benefits of a managed programming language like C\# (type safety, memory management, etc.), while targeting the demanding performance requirements of these applications.

Multimodal, integrative-AI applications are often extremely compute intensive and need to operate under strict latency constraints in order to act in the world in real time. The \psif runtime was designed to maximize resource utilization via multi-threaded pipeline parallelism, while insulating developers from the many challenges that often arise in concurrent execution environments. Automated data cloning ensures component isolation, and shared memory mechanisms and other optimizations minimize garbage collections when the pipeline is executing at a steady state.

Most interesting real-world systems will require a large amount of heterogeneous components. The \psif framework promotes a programming model where individual components are lightweight and are easily wired together to accomplish larger computation tasks. This model, in which components execute concurrently and in a coordinated way, enables efficiency gains via pipeline parallelism.

A central aspect of the \psif streaming infrastructure is that \emph{time} is a first-order construct. All messages flowing on streams carry not only a creation timestamp, but also an \emph{originating} timestamp which corresponds to the time when that message entered the pipeline from the real world. This timestamp is critical for proper synchronization and fusion across streams. \psif provides a large set of primitives for data synchronization and interpolation that a developer can invoke and configure, while hiding underlying complexities of concurrency, buffering, waiting for messages to arrive, etc. The \psif runtime empowers developers to optimize their pipelines with various fine-grained controls and throttling mechanisms for regulating flow on data streams.

\begin{figure*}[!t]
    \centering
    \includegraphics[width=\textwidth]{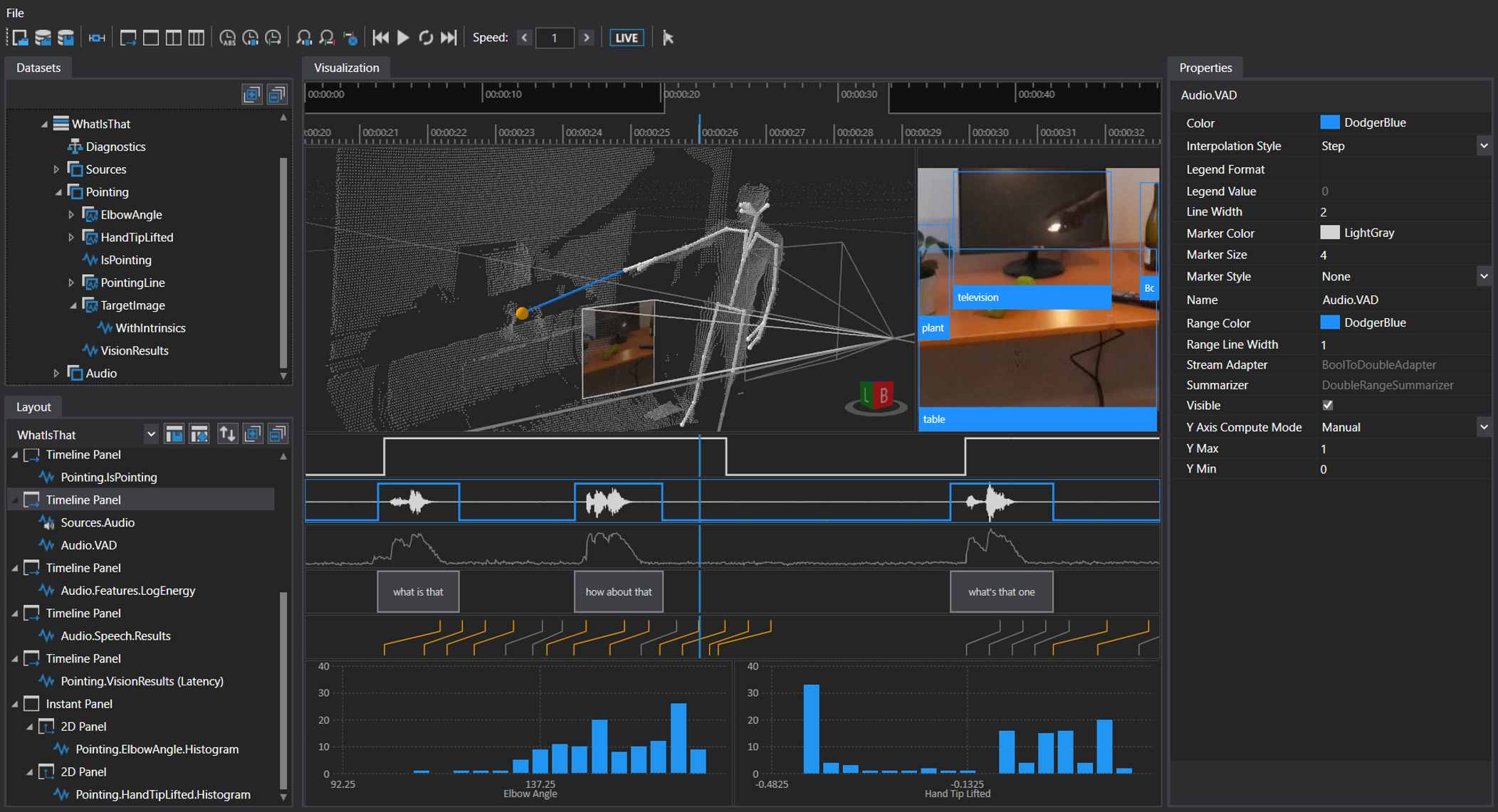}
    \caption{Data visualized in Platform for Situated Intelligence Studio, from a \psif application that identifies objects that a user is pointing to. The center panel shows a composite visualization layout that includes 3D, 2D, and timeline views.}
    \label{fig:PsiStudio}
\end{figure*}

Another important feature of the \psif runtime is that it helps speed up development by providing APIs for data persistence and replay. \psif allows developers to easily log streams of any data types to disk: the framework automatically generates performant serializers. While incrementally developing and debugging complex integrative applications, it is often time consuming and expensive to have to repeatedly run the application ``live.'' The persistence system enables developers to experiment with and execute pipelines where the input streams are read from disk, rather than produced by live sensors. Because of the fine-grained timing information logged in the data, pipeline executions can be  reproducible and deterministic.

\subsection{Debugging and Visualization Tools}

Platform for Situated Intelligence Studio, or \emph{PsiStudio} (Figure \ref{fig:PsiStudio}), is a data visualization and processing tool for the temporal, multimodal data generated by \psif applications. Users can construct and reuse complex visualization layouts, and they can easily navigate temporally through the data, inspect values, and select and play-back segments at varying speeds. The visualizations can be performed either offline, i.e., based on the data persisted to disk by the application, or live, while the application is running.

A wide array of configurable visualizers for different data types are available, from simple numerical streams, to audio, video, complex 3D objects, etc. A number of universal visualizers that operate over streams of any type are also defined, e.g., to visualize the latency or originating times of messages. These visualizers provide only a starting point. Developers can write their own visualizers, configure the tool to load them from external assemblies, and also use visualizers outside of PsiStudio, in their own applications.

Figure \ref{fig:PsiStudio} shows data visualized in PsiStudio from a sample \psif application that attempts to identify the objects that a user is pointing to. The top-left panel contains a 3D visualization, in which information from the depth map, body tracking, pointing direction, and pointing target streams is simultaneously shown. Next to it on the right, a 2D panel visualizes a stream containing the cropped image around the pointing target, as well as an overlaid visualizer showing the object detection results. Below that are five timeline panels showing (1) a boolean stream indicating when the user is pointing, (2) an audio stream overlaid with a boolean voice activity detection stream, (3) a log-energy stream, (4) speech recognition results, and (5) a latency visualizer that provides information about the originating-time and latency of each object detection result (this latter visualizer was configured to show latencies larger than 500ms in orange). Finally, the bottom row shows two histogram visualizers. PsiStudio also enables developers to visualize the application pipeline itself, which is often a critical ability in debugging data synchronization and starvation problems, discovering which components are causing slowdowns, etc. The developer can visually inspect the structure of the pipeline they constructed in code, with the various components and streams, and drill down hierarchically into sub-components.

PsiStudio supports temporal data annotation scenarios based on custom, user-defined annotation schemas. This is an important capability for data-driven work, iterative refinement, and tuning. The annotated data is itself persisted as a \psif stream, enabling a variety of semi-automatic data labeling and training scenarios.

Finally, PsiStudio provides access to data processing functionality. Multiple stores of data collected from the same application over time can be organized hierarchically into larger sessions and datasets. Exploratory analyses can be performed and new streams can be computed from existing data in batch mode, e.g., extracting acoustic features from all audio streams in a dataset. This data processing functionality is additionally available via a command line tool.

\subsection{Open Ecosystem of Components}

Apart from the runtime infrastructure and tools described above, the \psif framework also includes a set of components which promote encapsulation and reuse, and provide the basis for rapidly prototyping applications. The components currently available in the GitHub repository center around multimodal sensing and processing technologies. They include \emph{sensor} components for USB cameras, microphones, and depth sensors such as Microsoft's Azure Kinect and Intel's RealSense; audio and visual \emph{processing} components for speech recognition, language understanding, object detection, and body tracking; \emph{wrapper} components, for instance that enable running machine-learned models in ONNX format, or provide access to Azure Cognitive Services; etc.

Also included are a set of low-level, generic components for manipulating data streams of any type, called \emph{stream operators}. An array of basic operators provide time-related functionality, mathematical and statistical operations, aggregation, windowing, etc. Large-scale integrative-AI applications often require compositing many heterogeneous components, with input and output data types that may not align well out-of-the-box. \psif includes operators for writing simple adapters that \emph{transform} data into the proper shape when interfacing between such components. Another important category of stream operators facilitate stream \emph{fusion} and \emph{merging}, and they provide the basis for reproducible, correct synchronization across streams. Still other operators enable the \emph{dynamic} construction of computation graphs that can change their structure throughout the execution process, depending on the data flowing through the pipeline. Finally, APIs exist for hierarchically encapsulating a sub-graph of multiple components into a single \emph{composite} component.

Overall, the runtime, tools, and components provided by \psif streamline development efforts both at the level of the \emph{component writer}, and at the level of the \emph{application writer}. The same developer can alternate between both roles, utilizing both off-the-shelf components provided in the framework or available from the community in order to assemble their application, while also occasionally writing new components to target specific functions needed by their application. New components can be easily developed and added in a way that shields component authors from the intricacies of the concurrent, coordinated execution environment. We hope the existing set of components will grow into an even larger ecosystem through community contributions, further lowering the barrier to entry for developing multimodal integrative-AI applications.

\section{Implications for HRI}

The scope and use-cases of Platform for Situated Intelligence are broad: any application that processes streams of data, and where timing is important, can benefit from its programming models, primitives, and tools. Examples range from analyses of multimodal data all the way to open-world social robots. As such, we believe the framework is particularly well-suited for researchers in HRI.

Given the acute needs in this space, a number of related infrastructure and development tools have been developed over the years, such as IrisTK, SSI, MediaPipe, and ROS. IrisTK \cite{skantze2012iristk} is a Java-based framework that focuses on multiparty face-to-face interaction and social robotics; it provides a set of modules for perception and production, and formalizes the authoring of dialog control around Harel state-charts. The Social Signal Interpretation (SSI) framework \cite{wagner2013social} provides tools that enable synchronized recording, analyzing, and recognizing human behavior in real time. MediaPipe \cite{lugaresi2019mediapipe} enables developers to easily create custom perception pipelines as graphs of modular components, connecting sensors to arbitrary ML inference models and other media processing components.

Most familiar to the HRI community is Robot Operating System (ROS) \cite{Quigley09}, which provides a message-passing infrastructure, an open ecosystem of components, and a set of tools that simplify development of robotics applications. In ROS, each component executes in its own process, and a topology-aware central node handles connections. Since inter-process communication is more costly, this tends to lead to the development of coarser, monolithic components. By contrast, \psif applications tend to contain a large number of fine-grained components, all residing in the same multi-threaded process, with the \psif runtime efficiently handling message scheduling and communication. This architecture affords powerful features such as delivery policies, throttling, and back-pressure.

Each of these frameworks has beneficial attributes, and all have been successfully used in research. Distinguishing characteristics of \psif include its programming model which aims to leverage pipeline parallelism and encourages small, light-weight components that are executed in a concurrent yet coordinated fashion, its large set of pre-built primitives for reproducible synchronization and for reasoning about and manipulating temporal streaming data, and its debugging and visualization tools to speed up development.

Platform for Situated Intelligence is designed as an extensible framework, and can bridge to and integrate with other ecosystems such as ROS, Python, JavaScript, Unity, etc. One can construct hybrid systems that take advantage of different affordances in different frameworks, e.g., a mobile social robot that uses ROS nodes for navigation and \psif components for social perception. In addition to what is currently built-in, developers can write their own third-party visualizers and third-party data importers, e.g., for WAV files, mpeg videos, ROS bags, and so on. \psif applications can execute on a single machine, or can be distributed across multiple machines through remoting capabilities.


We plan to continue to extend the functionality and components available in \psifnospace, with a particular focus on embodied, physically-situated interaction. We are working on more components and tools for multimodal perception of the situated social context, controllers for generating utterances, intentions, and actions, and realizers for rendering those actions onto the behaviors of a social robot or virtual agent.

Given the challenges around multimodality, integrating multiple technologies, and handling time and data synchronization, it is no surprise that we have still not seen large breakthroughs in the space of complete end-to-end systems that can interact with people in the real world, like we have for the individual component technologies and research that can be conducted offline over large datasets. Enabling people to make breakthroughs on larger integrative systems is what \psif was built to achieve. Ultimately, we believe lowering the barrier to entry in this space will rest to a large degree on fostering a community of users and contributors, and creating a thriving ecosystem of reusable components. 

To learn more and get started with the framework, please see \textcolor{blue}{\url{https://github.com/microsoft/psi}} for walkthroughs, samples, and documentation. We invite anyone who is interested to help improve and evolve the platform, and we welcome contributions across the board: from simply using it and filing issues and bugs, to writing and releasing new components, to contributing new features or bug fixes.

\section{Acknowledgements}
Platform for Situated Intelligence is the work of a larger team of engineers and researchers. We would like to acknowledge the current members of the \psif team, which includes Ashley Feniello and Nick Saw, as well as past members: John Elliott, Don Gillett, Anne Loomis Thompson, Mihai Jalobeanu, and Patrick Sweeney. We would also like to thank Eric Horvitz for his contributions and support, as well as our early adopters for their feedback and suggestions.

\bibliographystyle{AAAI}
\bibliography{references.bib}

\end{document}